# Neurocomputational Mechanisms of Syntactic Transfer in Bilingual Sentence Production

Ahmet Yavuz Uluslu[1, 2], Elliot Murphy[3]

1. Linguistic Research Infrastructure, University of Zurich, Switzerland
2. Machine Intelligence Laboratory, Department of Engineering, University of Cambridge, UK
3. Vivian L. Smith Department of Neurosurgery, University of Texas Health Science Center at Houston, Texas, USA

**Corresponding author:** ahmetyavuz.uluslu@uzh.ch

**Abstract**: We discuss the benefits of incorporating oscillatory neural mechanisms into the study of bilingual production errors and their traditionally documented timing signatures (e.g., event-related potentials), which can offer new implementational-level constraints for theories of bilingualism. We argue that a recent neural model of language, ROSE, can offer a neurocomputational account of syntactic transfer in bilingual production, capturing some of its formal properties and the scope of morphosyntactic sequencing failure modes. We take as a case study cross-linguistic influence (CLI) and attendant theories of functional inhibition/competition, and present these as being driven by specific oscillatory failure modes during L2 sentence planning. We argue that modeling CLI in this way not only offers the kinds of linking hypotheses ROSE was built to encourage, but also licenses the exploration of more spatiotemporally complex biomarkers of non-native processing than more commonly discussed neural signatures.

## 1. Introduction

The coexistence of multiple languages in a single mind is a common feature of human cognition (indeed, bilingualism is much more common than monolingualism on a global scale) and has profound consequences for linguistic representation, production, and processing (Leivada et al., 2021; Sabourin & Stowe, 2008). While the structural adaptations and functional control mechanisms associated with second language acquisition are well established, the implications of bilingual co-activation for neurocomputational theories of language are less commonly addressed. Yet there are many aspects to the psychology of bilingualism that hold great relevance for testing hypotheses concerning how core properties of language are neurally enforced. In this article, we offer some cases that provide theoreticians and experimentalists with ways of further refining neural models of morphosyntax. (see Fig. 1)

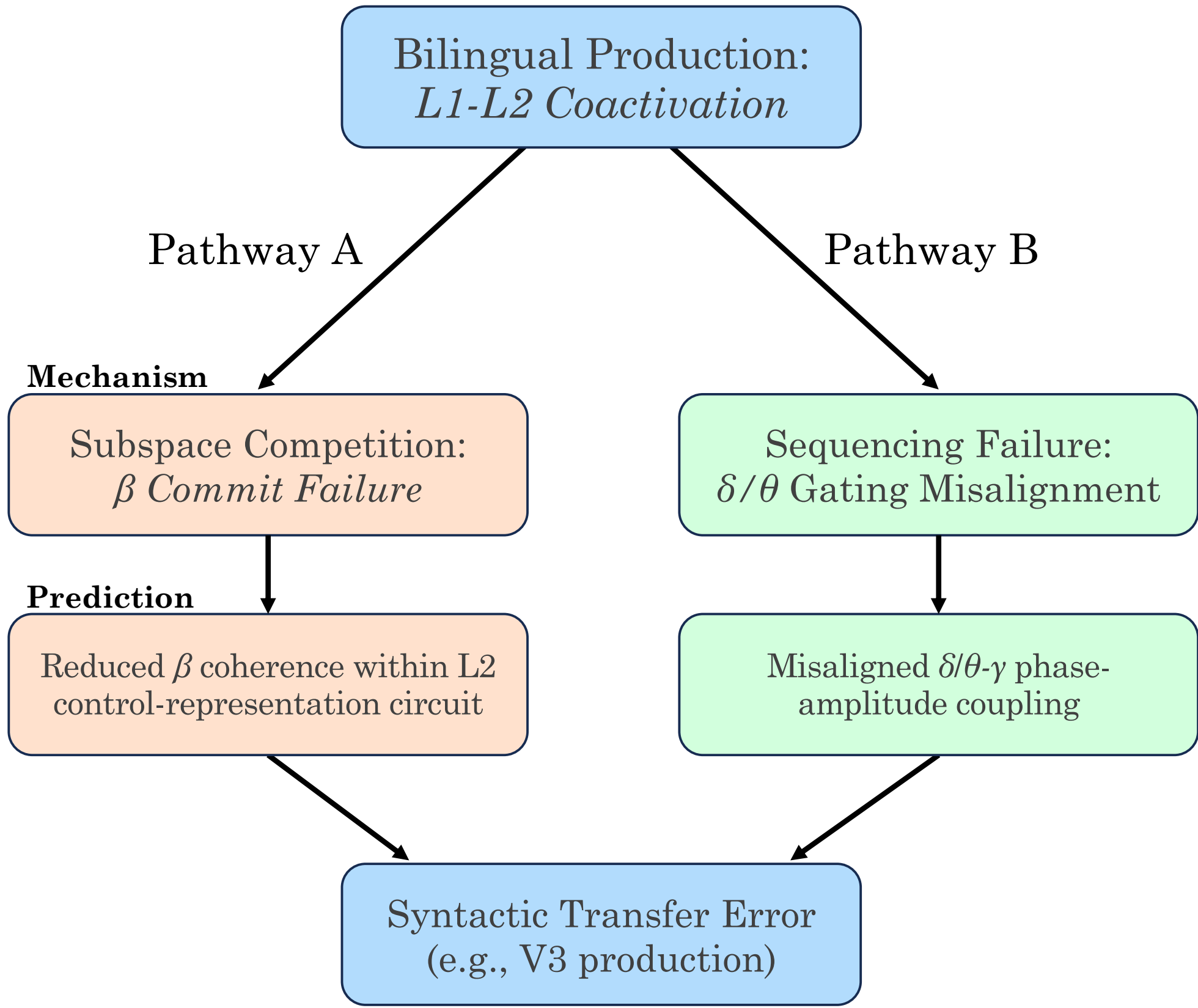

**Figure 1: Two pathways of syntactic transfer under ROSE**. Pathway A (Subspace Competition) describes non-native-like output driven by insufficient β-mediated stabilization of an intended L2 plan, allowing an entrenched L1 attractor to dominate. Pathway B (Sequencing Failure) describes errors driven by misaligned δ/θ phase-coded gating such that γ motifs are recruited at non-target phases, yielding word-order violations (e.g., V2-to-V3 shifts).

To pursue this goal, we will focus on Cross-Linguistic Influence (CLI). This refers to the influence of any language(s) a person knows on their learning, knowledge, or use of any other language(s) (Jarvis & Pavlenko, 2008). CLI is an inherently bidirectional and dynamic process, modulated by proficiency and language experience. The present article uses L1-to-L2 syntactic influence—alternatively referred to as syntactic transfer in the SLA literature—in bilingual sentence production as a case study. L1 influence is now understood to be a persistent and likely inevitable characteristic of bilingual cognition (Van Dijk et al., 2022). Even highly proficient speakers retain unconscious traces of their L1 that are detectable across linguistic levels (Markov et al., 2022), from phonology to semantics to morphosyntax. As a running example, we focus on non-native-like V3 production in the acquisition of verb-second (V2) word order. In V2 languages, a clause-initial adverbial or other non-subject constituent is typically followed by the finite verb in main declarative clauses (Vikner, 1995). Learners whose L1 lacks V2 are more likely to produce V3 sequences than learners whose L1 has V2, placing the subject before the finite verb in contexts where the target language requires verb-second order (Søby & Kristensen, 2025).

Current theoretical frameworks approach CLI primarily through functional levels of explanation and provide representational accounts of how language-specific structures compete for activation (Smith & Truscott, 2014). These models were not developed to explicitly address questions of neural dynamics; namely, *how* various psycholinguistic processes are neurally realized. Consequently, while we can consult rich descriptive accounts of *what* representations compete, we lack mechanistically explicit linking hypotheses for *how* the brain physically resolves this competition.

As such, we aim to target a major gap in contemporary L2 processing research: the lack of explicit neurocomputational accounts of how L1 interferes with sentence production in L2. Achieving this will allow researchers to go beyond high-level functional/computational notions of inhibition or resource allocation, towards metrics that deliver causal-mechanistic insight. To bridge this gap, we adopt the ROSE (Representation, Operation, Structure, Encoding) neurocomputational model (Murphy, 2024, 2025). By explicitly mapping symbolic linguistic operations onto testable oscillatory dynamics, ROSE offers a series of linking hypotheses concerning abstract linguistic structures and their implementational basis (see also Murphy, 2015, 2020). Using this framework, which posits that oscillatory control of cortical space constitutes a fundamental computational dimension (Chen et al., 2026), we propose that syntactic transfer can arise from two distinct neurocomputational mechanisms: Subspace Competition and Sequencing Failure. The former represents a commitment failure (via the β-mediated 'Commit' operation of ROSE) where the deeply entrenched L1 template prevents the L2 plan from achieving β power (13–30 Hz) stabilization. The latter represents a temporal mismatch where L1 oscillatory rhythms gate the sequential recruitment of motifs at the wrong moment.

ROSE decomposes syntax into four neurocomputational components. **R**epresentation is the coding of atomic lexico-semantic and morphosyntactic features at the single-unit and ensemble level. **O**peration is the transformation of these features into manipulable bundles, implemented through high-frequency broadband gamma activity. **S**tructure is the recursive, hierarchical organization of these bundles, carried by low-frequency synchronization and cross-frequency coupling. **E**ncoding is the externalization of structure into cortical workspaces via frontotemporal traveling waves. The single linking idea is that the *phase* of slow rhythms gates the high-frequency operations through phase-amplitude coupling (PAC): low-frequency phase opens excitability windows that license specific operations. Crucially, these dynamics recruit the brain's general-purpose machinery for maintenance, gating and binding. What is language-specific in ROSE is not the rhythms but their representational content, interactions, and task context: the features, bundles and plans that the dynamics coordinate during sentence production, and the algebraic properties they enforce.

Within the broader literature on oscillations and language, ROSE can be positioned against two families of accounts. Hierarchical-tracking accounts hold that low-frequency activity tracks and segments constituent structure (Ding et al., 2016), while compositional architecture accounts emphasize the construction and integration of structured representations (Coopmans et al., 2023). ROSE is distinctive in tying specific *phase-coded* operations to falsifiable production signatures: different phases of a slow cycle license distinct actions (e.g., retrieval vs. constituent reduction; Murphy, 2025). It is this property – a concrete, gating-based control scheme – that makes ROSE suited to predicting non-native-like word-order outputs, which are naturally construed as action-selection or gating events under co-activation (see Murphy, 2025 for extensive discussion of comparisons between ROSE and other accounts).

Concretely, our running case is the production of verb-second (V2) word order by learners whose first language lacks it. In V2 languages such as Danish, when a main declarative clause begins with a non-subject constituent (e.g., an adverbial), the finite verb must appear in second position, before the subject. Learners whose L1 is not V2 (e.g., English) are more likely to produce non-target V3 sequences, placing the subject before the finite verb; for example, producing *Nu jeg bor i Danmark* ('Now I live in Denmark') in place of the **target** V2 form *Nu bor jeg i Danmark* ('Now live I in Denmark'). We use this well-documented case (Søby & Kristensen, 2025) as a worked example throughout, while treating L1-to-L2 influence as one direction of an inherently bidirectional, proficiency- and input-graded phenomenon (Serratrice, 2012).

The remainder of this article is organized as follows. Section 2 reviews some functional accounts of bilingual syntax. Section 3 details the neurocomputational toolkit that we will subsequently recruit. Section 4 articulates the two proposed pathways of interference. Section 5 demonstrates the utility of this framework by analyzing well-documented syntactic influence phenomena in L2 production. Section 6 discusses limitations and future directions. Section 7 concludes.

## 2. Cross-Linguistic Influence

Having sketched the ROSE model, we now review the relevant bilingual production phenomena that any mechanistic account must explain (behavioral transfer and its neurophysiological signatures), before returning in Section 3 to specify the mechanism in full.

### *2.1. Behavioral Evidence*

Cross-Linguistic Influence (CLI) is defined as the influence of any language(s) a person knows on their learning, knowledge, or use of any other language(s) (Jarvis & Pavlenko, 2008). In the context of bilingualism, this means a speaker's L1 and L2 continually interact across all linguistic domains, extending from phonology to morphosyntax (Kroll et al., 2006; Marian & Spivey, 2003). This interaction can facilitate acquisition when structural properties align, or it can induce processing constraints when structures diverge, often resulting in non-native-like production. This structural influence is not an absolute mechanism, since both facilitation and interference effects are highly conditional, depending on factors such as the speaker's proficiency, the degree of structural overlap, and specific cognitive demands (e.g., Bialystok & Craik, 2022; Cunnings, 2017; de Bruin et al., 2023; Radman et al., 2021). Furthermore, whether such an integrated system should even be evaluated against a monolingual standard remains a subject of intense debate. Contemporary accounts suggest that the bilingual speaker operates within a unified, permeable space where languages are not strictly separated, but permanently coordinate and compete for activation within a shared linguistic subspace (Kroll et al., 2015).

We note here that we are using three notions that are often run together and should be kept distinct. *Co-activation* is the empirical fact that a bilingual's languages are simultaneously active during processing (Marian & Spivey, 2003). *Language control* refers to the mechanisms that bias selection toward the target language and suppress the non-target one (Green, 1998). *Cross-linguistic influence (CLI)* is the structural or representational *outcome* – one language's structure shaping another's – that can arise when co-activation is incompletely resolved. Co-activation is thus a precondition for CLI rather than CLI itself, and the two pathways we propose below are candidate implementational mechanisms for how unresolved competition becomes a structural outcome.

The foundation of this shared subspace is the persistent co-activation of multiple languages within a single brain (Dijkstra, 2005). Because representations from both languages constantly vie for selection, the system relies on cognitive control mechanisms (widely argued to recruit domain-general executive networks) to resolve cross-language conflict and suppress non-target interference (Bellegarda & Macizo, 2021; Blumenfeld & Marian, 2014; Green, 1998; Kroll et al., 2006; Luque & Morgan-Short, 2021). Lexical competition is not itself peculiar to bilinguals: within-language competition from phonological neighbors is well documented in monolingual processing (Vitevitch, 2002). What is distinctive about bilingual processing is the addition of cross-language competitors: candidates from the non-target language that remain active and compete for selection (Colomé & Miozzo, 2010; Marian & Spivey, 2003). For example, in picture-naming tasks, phonological neighbors in non-target languages have been shown to create significant interference, modulating reaction times and accuracy (Colomé & Miozzo, 2010). Similarly, eye-tracking data reveal that bilinguals frequently fixate on distractors with phonological overlap in the suppressed language, confirming that the non-target lexicon remains highly active during processing (Marian & Spivey, 2003).

Crucially, the mechanisms governing this co-activation and competition are not limited to isolated word production (e.g., in single-word picture naming tasks). The same alignment and competition mechanisms operate along a single linearization continuum, from sub-word morphosyntactic assembly (e.g., verb inflection, tense, aspect) to multi-word sentence production (Krauska & Lau, 2023). Framing the relevant unit as the 'word' would mis-describe agglutinative and polysynthetic languages, where the linearized units are sub-lexical; we therefore treat feature bundles, not words, as the operative unit throughout.

Consequently, when grammatical structures align, established L1 structures can boost performance, as the speaker's predictive, generative model gains support from multiple sources (Friston et al., 2020; Murphy et al., 2024). This facilitation is driven by a shared-syntax architecture, where bilinguals utilize the same abstract combinatorial nodes across languages for structurally similar constructions (Hartsuiker et al., 2004). Cross-linguistic structural priming studies, for instance, demonstrate that bilinguals are significantly more likely to reuse a specific syntactic structure across languages when the L1 and L2 share the same constituent word order (Bernolet et al., 2007). However, the ability to utilize these shared structural representations is heavily modulated by L2 proficiency (Hartsuiker & Bernolet, 2017).

Conversely, when structures conflict, or when L2 structural nodes are not yet fully developed, the entrenched L1 structure is significantly more likely to intrude upon the L2 sentence plan. This interference can induce processing slowdowns (Hopp, 2016) or structural deviations (Hsin et al., 2013). Recent analyses of learner texts confirm that this phenomenon is highly dependent on both L1 background and sentence complexity. The clearest recent quantification comes from Søby and Kristensen (2025), who analyzed L2 Danish written production across learners from 52 first-language backgrounds. The design contrasted learners whose L1 is V2 (e.g., German, Dutch) with learners whose L1 is not (e.g., English, Romance languages). The key finding was a group difference in non-target V3 production in obligatory-V2 declarative contexts. Non-V2-L1 learners produced V3 order in 115 of 428 relevant clauses (26.9%), for example producing the L1-congruent sentence *Nu jeg bor i Danmark*, with the subject-before-verb, in place of target V2 *Nu bor jeg i Danmark*. In comparison, V2-L1 learners produced V3 order in 6 of 63 relevant clauses (9.5%). For non-V2-L1 learners, target V2 production also increased across test levels. The test-level effect was marginal in the full model but significant in the post-hoc analysis restricted to non-V2-L1 learners. Critically, non-target order was more frequent when the clause began with a heavy adverbial phrase, and when the subject itself was multi-word rather than a single pronoun.

We note that these transfer effects are not an all-or-nothing 'template transfer'. Both facilitation and interference are highly conditional on proficiency, the degree of structural overlap, and task demands. We return to each of these as a mechanistic variable rather than a caveat: under ROSE, proficiency scales the β amplitude available to stabilize an L2 plan, structural overlap determines whether a shared motif can be recruited, and load and time pressure determine which of the two failure pathways dominates.

To explain the mechanisms underlying these behavioral outcomes across the linearization continuum, computational accounts of CLI frame transfer as a consequence of competition between partially activated structural candidates. Modular processing frameworks such as MOGUL (Smith & Truscott, 2014) treat bilingual production as the outcome of interactions among representational systems and the control mechanisms that regulate which language-specific structures are selected for output. In this view, L1 influence arises when selection is biased toward the entrenched L1 structure, either because inhibitory control is insufficient or because the L1 structure possesses a higher resting level of activation. While these frameworks provide rich functional descriptions of what competes, our present goal is to add an implementational hypothesis for how such competition is realized in neural time via oscillatory stabilization and phase-based gating. Below, we turn to some commonly explored neural signatures of CLI to mechanistically ground this phenomenon.

### *2.2. Neurophysiological Evidence*

While behavioral studies confirm the existence of L1 transfer, event-related potentials (ERPs) uncovered from scalp electroencephalographic (EEG) research provide the millisecond-level resolution necessary to distinguish underlying operations (Payne et al., 2020). In native processing, distinct neural signatures typically dissociate semantic from syntactic violations: the N400 (peaking ~400ms) indexes lexico-semantic integration (Kutas & Federmeier, 2011), while the P600 (peaking ~600ms) reflects structural reanalysis (Kaan et al., 2000; Osterhout & Holcomb, 1992). However, recent work challenges the uniformity of this profile, noting that even native speakers exhibit significant variability based on individual processing strategies (Freunberger et al., 2022; Sayehli et al., 2022). In non-native processing, this variability manifests as a qualitative shift in how violations are handled. Novice learners frequently exhibit N400 responses to syntactic violations that would trigger a P600 in native speakers (Mickan & Lemhöfer, 2020). This substitution suggests that beginners process structural violations as lexical anomalies, relying on associative heuristics rather than grammatical parsing. Cross-sectional evidence suggests that as proficiency increases, the N400 fades and is replaced by a P600, marking a transition from associative chunking to structural organization. However, the attainment of this native-like P600 is constrained by L1-L2 distance. When L2 structures directly conflict with the L1 template (e.g., V2 word order conflicts), even advanced learners often display a delayed P600, despite high behavioral accuracy (Andersson et al., 2019). Much of this neurophysiological evidence comes from comprehension rather than production. We treat it as relevant because it constrains the temporal signatures of L2 morphosyntactic processing.

While ERPs successfully map the timeline of these processing failures, they are limited as standalone explanatory mechanisms that might unveil how the algebraic properties of language are neurally enforced (Benítez-Burraco & Murphy, 2016, 2019). The ERP methodology relies on averaging signals across trials, a process that captures phase-locked potentials but systematically cancels out induced oscillatory activity (Rossi et al., 2023). This is critical because L2 processing is inherently variable, involving significant latency jitter as the system struggles to resolve competition. Crucially, an accumulating body of evidence suggests that key features of ERP components themselves may arise from stimulus- or task-induced phase resetting of ongoing low-frequency oscillations, rather than from additive, modality-specific evoked responses (see Lakatos et al., 2020; Murphy, 2020). ERPs may index *when* neural systems are perturbed, but they obscure the ongoing, intrinsic brain states (such as L1-tuned $\alpha/\beta$ rhythms) that condition those perturbations. Consequently, to understand how the brain physically mediates between L1 and L2 templates, we must look beyond averaged potentials to the induced oscillatory dynamics that appear to maintain states and bind features (Murphy, 2024).

### *2.3. Dynamical Motifs*

To explain the interference of L1 syntax in sentence production, we must specify the nature of the entrenched template itself. While traditional accounts view templates as symbolic rules or probabilistic tendencies, the neurocomputational perspective we wish to promote here requires a more specific definition. We propose treating such a template as a sequence-based attractor, a preferred trajectory through neural state space where the transition between linguistic elements is dynamically stable. Because the L1 pathway is highly reinforced, the system naturally gravitates toward this low-energy path. Drawing on themes explored by Labancová and Kazanina (2025), we consider dynamical motifs (DMs) as stable patterns of neural activity that implement specific factorized computations (such as assembling bundles of morphosyntactic features) within independent neural subspaces. In this sense, a DM is not a stored sentence frame such as Subject-Verb-Object, nor is it simply a word. Rather, it is a reusable local activity pattern supporting operations such as assembling a finite-verb feature bundle or preparing a subject-DP bundle. The ordering of such motifs into a V2, SVO, or other trajectory is imposed by slower structural dynamics. Given that syntactic operations require the integration of storage and control, we characterize these motifs not as locally confined representations, but as distributed frontotemporal loops (postulated to rely on dorsal stream tracts such as the arcuate fasciculus), in accord with contemporary theories concerning the neurobiology of language production (Yeaton, 2025). In this view, the motif is sustained by some component of oscillatory synchronization between inferior frontal control regions and posterior temporal representations. This explicit link between distributed neural dynamics and syntactic operations forms the core of the ROSE architecture (Murphy, 2024, 2025). In this framework, successful L2 production depends on the correct sequential recruitment of these distributed motifs, which serve to implement the bundling and sequencing of the most commonly generated (hence, emerging into motifs) lexico-semantic and morphosyntactic features. To find the acceptable word order in L2, the system must engage the specific DM responsible for the target operation (e.g., binding the Subject role, or coordinating the sequencing of one feature bundle before another) at the precise moment required by the L2 schema. For example, in producing a Danish V2 clause, a finite-verb motif may bundle tense, lexical-semantic, and phonological information associated with the verb, while a subject-DP motif may bundle nominal and thematic-role information. Native V2 production requires the finite-verb bundle to be recruited immediately after the initial constituent.

To make this concrete, consider the target Danish sentence '*i dag jagtede hunden katte*' ('today the dog chased cats'). A dynamical motif here is not the sentence frame and not a word, but is rather a reusable local activity pattern that performs one operation of bundling commonly co-occurring lexico-semantic features (Labancová & Kazanina, 2025; Murphy, 2025). A finite-verb motif bundles the features needed to realize '*jagtede*' – tense, agreement, and the verb's proprietary lexical-semantic content – into a single manipulable object. A subject-DP motif bundles the nominal and thematic-role features for '*hunden*' ('the dog', definite, singular, Agent). The adverbial '*i dag*' ('today') is itself a single feature bundle at this level even though it is later externalized as two phonological words. What makes the output V2 rather than V3 is not stored in these motifs: their ordering into a V2 trajectory is imposed by the S/E levels, which open a gating window for the finite-verb motif *immediately after* the initial adverbial. The motif supplies the content and the slow rhythm decides when it is released. This division of labor is what allows the same motifs to participate in V2 and non-V2 orders, and it is where the two failure pathways can intervene.

Under this lens, syntactic transfer can be modeled as reduced stabilization or mistimed recruitment of the target frontotemporal configuration under the influence of a more entrenched L1-compatible trajectory. The next section will outline further specifics about the neural model of morphosyntax that we argue can serve to better ground the psychology of bilingualism in a neurobiologically plausible basis.

Because motifs are carved by experience, proficiency enters the model as a continuous quantity rather than a switch between a 'wrong' and a 'correct' grammar. Two distinct sources of non-target output follow. The first is *incomplete acquisition*: under limited input the relevant L2 motif or atomic feature may not yet be installed at R/O, so there is no target plan to stabilize. The second is *competition or gating failure*: the target plan exists but loses the stabilization (Pathway A) or timing (Pathway B) competition. As input accumulates, the L2 attractor basin deepens, the β amplitude and duration available for a successful Commit increase, and the L2 low-frequency gating schedule becomes more reliable. The model therefore predicts a developmental ordering: low-proficiency learners will presumably show both incomplete-acquisition failures and frequent competition failures; high-proficiency learners largely will resolve the fast gating-misalignment errors of Pathway B but can retain residual Subspace Competition (Pathway A) failures under high maintenance load. This competition may explain why advanced learners still produce occasional V3 forms with long or structurally demanding clause-initial constituents, where the L2 plan must be maintained across an extended pre-verbal window. It may also be reflected in delayed P600 responses in advanced learners, although the evidence for the latter comes primarily from comprehension.

## 3. Neural Dynamics of Language

Alongside functional models which describe competition at the level of symbolic activations, we can also consider neural models that attempt to explain the formal and implementational nature of such symbolic forms of cognition. While previous researchers have successfully linked neural oscillatory activity to periods of language production (Howard & Poeppel, 2012; Piai et al., 2014), the ROSE model (Murphy, 2024, 2025) provides explicit linking hypotheses between syntactic structures and rhythmic brain signatures. In this section, we will position ROSE not as a replacement for higher-order functional theories like MOGUL (Smith & Truscott, 2014) but as a complementary mechanistic implementation.

While functional models successfully characterize *what* competes (e.g., L1 vs. L2 lemmas), ROSE specifies *how* this competition plays out in the physical constraints of neural time and space (Murphy, 2024, 2025). The framework decomposes syntax into four distinct neurocomputational components: **R**epresentations, **O**perations, **S**tructure, and **E**ncoding. Central to ROSE is the premise that the brain does not process language as a static sequence, or even a hierarchy, but as a dynamic trajectory of neural states governed by specific frequency bands. Certain properties of neural oscillations, pertaining to phase and amplitude coupling dynamics (Buzsáki, 2006; Canolty et al., 2006; Hyafil et al., 2015), are argued under ROSE to comply with known algebraic aspects of syntax (Murphy, 2025: 64-69). This allows for a functional dissociation between the neural rhythms responsible for atomic representational content (and 'Operations' over this space) and those responsible for structure (and the 'Encoding' of such structure in cortical workspaces). These mappings should not be understood as claims that particular frequency bands are dedicated exclusively to language. β stabilization, α inhibition, δ/θ timing, and γ-mediated local processing are of course widely implicated in domain-general processes such as working memory, attentional control, prediction, and sequential action planning. The specificity of the present proposal lies in the representational content and task context: these domain-general dynamics are hypothesized to coordinate morphosyntactic features and production plans during sentence generation.

Below, we briefly outline elements of ROSE by way of commonly discussed frequency bands that will pertain most saliently to properties of bilingual sentence production under discussion here.

- *Gamma* (γ, ~60–150 Hz): **Operations and Motifs**
  The O level of ROSE handles the transformation of atomic features into manipulable linguistic objects. These operations are implemented via high-frequency broadband γ activity (Fernandez-Ruiz et al., 2023; Wang et al., 2012). In our framework (as in Murphy, 2025), we align this level with DMs and the stable, factorized patterns of neural activity that encode specific computations (such as assigning a thematic role by virtue of triggering a bundle of lexico-semantic features) within independent neural subspaces. Each motif operates within a high-frequency γ burst, allowing for rapid feature binding within local circuits.
- *Beta* (β, 13–30 Hz), *Alpha* (α, 8–12 Hz): **'Commitment' and 'Shielding'**
  While γ handles rapid operations, the β band is responsible for the maintenance of the current neural state over time, in addition to rapid transitions in such states, e.g., syntactic-semantic anticipation (Engel & Fries, 2010; Murphy et al., 2022). β synchronization serves what is termed in Murphy (2025) a 'Commit' operation, stabilizing the current structural hypothesis. Future experimental work should aim to refine what properties of linguistic structure are maintained here, but evidence reviewed elsewhere (Murphy, 2020, 2024) suggests that at least some lexico-semantic and syntactic features are maintained until they are resolved. $\alpha$ inhibition, meanwhile, has been argued to 'shield' this ongoing representational space from decay and interference (Murphy, 2024). This stabilization is crucial for processing longer constituents and dependencies where neural states must be sustained over time.

- Delta/Theta ($\delta$/$\theta$, <8 Hz): **Compositional Structure**

  The S level is responsible for the recursive or hierarchical organization of these objects. ROSE posits that this structure building is not static but relies on low-frequency synchronization and cross-frequency coupling. Slower rhythms, specifically in the $\theta$ (4–8 Hz) and $\delta$ (1–4 Hz) bands, provide temporal segmentation into which the faster motifs are coordinated. But since this 'chunking' alone is not a candidate mechanism for natural language syntax, Murphy (2025) shows how algebraic properties of syntax (commutativity, non-associativity) can be enforced via specific configurations and mathematical properties of phase-amplitude coupling being driven by these slower rhythms. Ideally, these slow rhythms coordinate the distributed frontotemporal loops necessary for syntactic processing.

Moving to a space already introduced above, under ROSE DMs are not themselves the carriers of hierarchical structure (Murphy, 2025). Rather, they are the reusable computational microcircuits that implement feature-bundling and locally factorized operations at the R/O levels. The hierarchical and linearization constraints that define a target L2 plan are imposed by slow phase codes at the S/E interface, which gate when (and which) motifs can be recruited. This division of labor follows ROSE's prediction that motifs should not directly encode syntactic variables such as node depth or address, but instead become organized into syntactic trajectories by low-frequency timing and cross-frequency coupling.

The final component of ROSE, Encoding, governs the externalization or linearization of this hierarchy into workspaces. In this architecture, successful sentence production is defined by a precise temporal synchronization between the structural and operational levels via phase-amplitude coupling (PAC). During production, the phase of the slow structural rhythm acts as a gating mechanism. It opens specific excitability windows that allow the recruitment of high-frequency motifs at the correct moment in time. For example, in a V2 language such as German, the L1 $\theta$ rhythm creates a specific temporal slot for the verb immediately following the first constituent.

We propose that L1 interference is a mechanistic failure of this coupling: either the L1's slow structural rhythm forces the L2 content into the wrong temporal slot (*Sequencing Failure*), or the L1's deeply entrenched gamma motifs compete for the same neural subspace which would percolate 'up' the ROSE hierarchy to influence the inter-areal feature workspace (*Subspace Competition*). These disruptions mirror the two core properties of PAC. The following sections will elaborate on these proposals.

### *3.1. Producing V2 in Danish*

To show the architecture working end-to-end, we trace here the production of '*i dag jagtede hunden katte*' ('today the dog chased cats') through the four components of ROSE.

**Representation (R)**. The conceptual message activates the atomic lexico-semantic and morphosyntactic features of all elements: TODAY (temporal adverbial); DEFINITE, SINGULAR, dog, AGENT; CHASE, PAST, 3rd-person; cat, PLURAL, INDEFINITE, THEME. At this level, these are distributed feature codes hosted by mixed-selectivity cell ensembles, not words or positions.

**Operation (O)**. Gamma-mediated activity assembles these features into a small number of manipulable objects, some of which would be subject to dynamical motifs that coordinate trajectories of firing rates for commonly co-occuring feature sets. The relevant unit here is the *feature bundle*, not the orthographic word: TODAY becomes a single adverbial bundle even though it will surface as two words (*i dag*); the verb features become a single finite-verb bundle (*jagtede*); the subject features become a subject-DP bundle (*hunden*); the object features a further DP bundle (*katte*). The model thereby generalizes to languages where the relevant units are sub-lexical.

**Structure (S)**. Low-frequency synchronization and cross-frequency coupling impose the hierarchical relations and the V2 ordering constraint over these bundles.

The slow rhythm establishes the temporal grid for the clause. In a V2 grammar it opens an excitability window for the finite-verb bundle immediately after the clause-initial bundle. A beta-mediated Commit burst stabilizes this structural hypothesis.

**Encoding (E)**. Frontotemporal traveling waves help externalize the committed structure into the production workspace, and PAC gates the release of each bundle at its licensed phase: the adverbial bundle is read out first, then (at the next licensed phase) the finite-verb bundle, then the subject and object bundles, with E linearizing the single TODAY bundle into the two surface words *i dag*. The result is the V2 string *i dag jagtede hunden katte*.

Two points of note here prepare the bilingual case. First, the ordering information lives in the slow S/E timing, not in the motifs, so the same motifs could be released in a non-V2 order if a different rhythm gated them. Second, two things must both succeed: the structural hypothesis must be *stabilized* (a strong enough β Commit) and each bundle must be *released at the right phase* (correct low frequency gating). The two failure pathways we will now discuss follow directly from these observations.

## 4. Two Pathways of Interference

With elements of ROSE able to capture properties of the morphosyntactic linearization process, we can now simulate how this system behaves under the stress of bilingual co-activation. We propose that cross-linguistic influence is not a monolithic failure of inhibition, but manifests through two distinct neurocomputational pathways: Subspace Competition (a failure of stabilization) and Sequencing Failure (a failure of temporal gating).

### *4.1. Pathway A: Subspace Competition (β-Mediated 'Commit' Failure)*

One possibility is that the language system at the R/O levels experiences subspace competition: namely, through conflicting DMs and 'Commit' failure. This first pathway concerns the stability of the linguistic representational plan; or, "which lexico-semantic features are being bundled into accessible syntactic objects?" Under ROSE, the maintenance of a structural state, such as holding a Subject-related motif active while planning the verb, is governed by widespread β power increases over relevant frontotemporal language sites recruited for morphosyntactic control and attention. This β activity serves to commit or stabilize the current internal model against noise, maintaining the relevant neural subspace active. Notice here that the 'subspace' of competition is not strictly neuroanatomical, but representational and spectral. We might formulate a subspace more precisely as an ensemble-defined feature space at the R/O levels, measured via consistent spectrotemporal patterns and connectivity motifs in high-frequency $\gamma$ activity, where this activity reports the cellular substrates, communication channels and computational operations underlying information processing in a given neural circuit (Fernandez-Ruiz et al., 2023).

In L2 production, we argue that interference arises as a competition for β-based stabilization. Because the L1 attractor landscape has been shaped by years of experience-dependent tuning, L1 templates occupy more regularly (and more deeply) carved valleys in neural state space. Consequently, when an L2 speaker attempts to instantiate a target structure, the competing L1 template is easier for β Commit operations to influence. The non-native-like production occurs not because the L2 rule is unknown, but because the L2 plan fails to generate sufficiently powerful β burst amplitude to override the entrenched L1 attractor. The result is a Commit failure where the neural subspace effectively defaults to the L1 configuration, as the system follows the path of highest stability. Critically, while L2 learning does sculpt new attractor basins, these are constrained by earlier-established temporal coordination regimes, making sequencing failures persist although they are expected to decline with increasing proficiency.

Bilingual language processing typically recruits largely overlapping frontotemporal networks. As noted already, the relevant distinction here is representational and dynamical: overlapping populations may instantiate different language-compatible trajectories depending on current connectivity, oscillatory state, task demands, and feature-decoding profile. On this view, L1-L2 competition should not be understood as competition between separate cortical territories, but as competition between partially overlapping neural trajectories within a shared language network. The critical prediction is therefore that, during non-native-like L2 production, the intended L2-compatible trajectory will show weaker β-mediated stabilization than the competing L1-compatible trajectory. This may be observable as reduced β burst amplitude, reduced β burst duration, or reduced β-mediated frontotemporal coherence in the task-relevant planning network, rather than as lower β power in anatomically L2-specific regions. Hypothesis 1 summarizes this:

> **Hypothesis 1** (β-Stabilization Hypothesis)
>
> During L2 sentence production, non-native-like output will be preceded by reduced β stabilization of the intended L2-compatible plan *relative to the competing L1-compatible plan within the same, shared frontotemporal network*, observable as reduced β-burst amplitude, reduced β-burst duration, and/or reduced β-mediated frontotemporal coherence in the task-relevant planning circuit, and *not* as lower β power in anatomically L2-specific regions.

The effect is predicted to be *interactional*: β destabilization should predict non-native-like production specifically when production difficulty co-occurs with L1-L2 structural conflict, and should be absent (or much reduced) in matched lexical retrieval or semantic search difficulty conditions that lack structural conflict. This prediction assumes partially overlapping neural populations whose current oscillatory state, connectivity profile, and feature-decoding pattern support either an L1-compatible or L2-compatible production trajectory. Crucially, β disruption should be strongest when production difficulty is combined with L1-L2 structural conflict. β dynamics are also sensitive to working memory, attention, prediction, and general production difficulty. β destabilization should be especially predictive of non-native-like production when L1 and L2 impose competing structural trajectories. If reduced β merely reflects general difficulty, comparable reductions should occur in matched lexical or semantic difficulty conditions without L1-L2 structural conflict. If the present account is correct, β disruption should be strongest when production difficulty is combined with structural competition. Thus, the relevant effect is not a global reduction in β activity, but a selective reduction in β stabilization within the task-relevant planning circuit supporting the target L2-compatible trajectory.

The interaction called upon in Hypothesis 1 is what distinguishes structural-template competition from generic production difficulty. β dynamics are domain-general and are modulated by working memory, attention, prediction, and difficulty; the present claim is not that reduced beta indexes syntax per se, but that reduced beta stabilization of the *target structural trajectory*, under structural conflict specifically, indexes the failure to Commit an L2-compatible plan. A design that crosses structural conflict (V2 vs. non-conflicting order) with matched lexical/semantic load therefore provides the critical, falsifying control: if reduced β merely reflected difficulty, comparable reductions should appear in the no-conflict load conditions.

Lastly, we do not assume here that there need to exist cortical sites preferentially devoted to L2 over L1 features. Current evidence indicates that a bilingual's languages recruit largely overlapping frontotemporal networks (Sulpizio et al. 2020). The competition we describe is therefore *representational and spectral, not anatomical*. This construal has a direct methodological consequence that sharpens Hypothesis 1. Because the competing L1- and L2-compatible plans are instantiated by overlapping populations, the prediction cannot be tested by contrasting β power across anatomically defined L1 and L2 'regions': no such regional dissociation is expected, and finding one would, if anything, count against the present account. The operative quantity is defined at the level of the trajectory rather than the cortical site. What must be recovered is the population activity pattern corresponding to an L2-compatible plan, together with the β-band stabilization of that pattern across the planning window. This requires trajectory-level decoding rather than local spectral power, and two complementary approaches are available. First, state-space methods (e.g., targeted dimensionality reduction or demixed PCA) can isolate the low-dimensional subspaces along which the competing plans evolve, allowing β-burst amplitude and duration to be quantified for each trajectory separately and on a trial-wise basis (Kobak et al., 2016; Mante et al., 2013) – the same subspace logic that defines dynamical motifs at R/O (Labancová & Kazanina, 2025). Second, connectivity-motif measures (β-mediated frontotemporal coherence and directed connectivity within the task-relevant planning circuit) index the stabilization of a distributed configuration rather than power at any single recording site. On this operationalization, the diagnostic observation is reduced β stabilization of the *decoded* L2-compatible subspace under structural conflict, with local regional β power held constant; its absence would falsify the pathway.

### *4.2. Pathway B: Sequencing Failure (δ/θ Gating Misalignment)*

A second possibility for morphosyntactic interference effects is fundamentally temporal in nature. As operationalized above, linearization relies on PAC, where slow structural rhythms open gating windows for content. We propose that syntactic transfer patterns, such as the V2 violations observed in L2 Danish, represent a failure of this temporal segmentation. This occurs when the L1 chunking window remains dominant during L2 production. In this scenario, the L1's low-frequency prior imposes its own segmentation onto the L2 stream, opening gating windows at intervals consistent with L1 syntax but incompatible with the L2 target. For example, if the L1 rhythm gates the Subject slot immediately after an initial adverb, the L2 Subject feature bundle will be triggered prematurely, regardless of the speaker's declarative knowledge of the L2 word order. In this instance, this L2 feature bundle may or may not be represented as a DM, depending on the proficiency of the language user. This is a sequencing failure where the correct motifs are recruited, but they are slotted into a temporal grid generated by the wrong oscillator.

We are careful here not to claim a categorical 'slot' for each syntactic position. Rather, low-frequency *phase* sets excitability windows that bias which bundle is read out, the mechanism for which there is direct human evidence (phase-amplitude coupling supporting phase coding in human cortex; see Murphy, 2025 for review). This is how our proposal relates to hierarchical-tracking accounts discussed above: those accounts establish that low-frequency activity *segments* constituent structure; ROSE adds that the *phase* of that activity gates which operation is licensed (Murphy, 2025, Fig. 5), so a non-native-like V3 arises when the subject bundle is read out at the phase that, in the target grammar, is reserved for the verb. As with Pathway A, the coupling machinery is domain-general; what is language-specific is that it is hypothesized here to carry morphosyntactic gating during sentence planning. Hypothesis 2 summarizes this:

> **Hypothesis 2** (PAC Sequencing)
>
> Sequencing-related transfer patterns (e.g., V3 placement) will be associated with misaligned $\delta/\theta$-$\gamma$ phase-amplitude coupling, such that high-frequency motifs at R/O are recruited at non-target phases of the slow structural rhythm at S (e.g., $\gamma$ bursts occur earlier in the $\delta/\theta$ cycle on non-target V3 productions).

The prediction is, again, interactional: phase misalignment should track non-native-like order specifically under L1-L2 word-order conflict, not under matched sequencing demands without conflict. We expect that Pathway A is more likely to occur with longer or structurally demanding constituents, high working memory load, low speaker proficiency, and potentially also dual-task or noisy conditions. In contrast, Pathway B is more likely to occur with strong word order conflict (V2/V3), time pressure, and relatively fast production. For both hypotheses, a familiar combination of metrics will be relevant; trial-wise PAC analyses (e.g., modulation index) (Scherer et al., 2022), β burst metrics (rate/amplitude/duration) and coherence measures (Thatcher, 2012), in addition to the methods outlined in the previous section.

Overall, these pathways afford the functional constructs of bilingual production models a concrete implementational reading. *Degree of co-activation* corresponds to the relative depth of the competing L1 and L2 attractor basins; *inhibitory/control demand* corresponds to the β amplitude required for a successful Commit of the target plan; *similarity-based interference* corresponds to overlap in the gamma feature-bundles recruited by the two languages (high overlap yielding facilitation, partial overlap yielding the strongest competition); and *graded proficiency* corresponds to the progressive carving of L2 motifs and the increasing reliability of L2 low-frequency gating. ROSE does not – and is not intended to – displace functional accounts such as MOGUL. Rather, it supplies the implementational layer that such models were not designed to specify, translating what competes into how, and when, competition is resolved in neural time.

### *4.3. Facilitation Effects*

While L1-L2 co-activation can induce interference, it simultaneously provides the neurocomputational basis for facilitation. Under ROSE, structural facilitation (such as an L1 Spanish speaker efficiently acquiring L2 French grammatical gender) arises from shared or highly sympathetic dynamical motifs. Because both languages utilize similar feature-bundling operations, the L2 can recruit the pre-existing, deeply entrenched γ-band motif at the Operation level, significantly reducing the β-band amplitude required for Commit stabilization. Crucially, speakers do not systematically over-apply this gender motif to an L2 lacking it (e.g., English) because the target L2 representations lack the necessary atomic features at the R level to trigger the motif, and the L2 δ/θ structural rhythms do not open a gating window for its temporal insertion.

To put this another way, the same machinery that produces interference produces facilitation when languages align. Consider an L1-Spanish speaker acquiring grammatical gender in L2 French. Because both languages perform the same gender feature-bundling operation, the L2 can recruit a pre-existing, deeply entrenched γ-band motif rather than building one from scratch. A shared, highly available motif raises the activation of the target bundle and lowers the β amplitude required for a successful Commit. In effect, facilitation is mechanistically a cheaper Commit.

This account also explains why facilitation does not become over-application. Speakers do not systematically impose gender on a gender-less L2 (e.g., English) for two independent reasons. First, the target English lexical representations lack the atomic gender feature at R that would trigger the motif, so there is nothing to bundle. Second, even where a motif exists, the L2's low frequency structure does not open a gating window for its temporal insertion. Possessing a motif is therefore necessary but not sufficient: it must also be recruited by the L2 rhythm. Facilitation requires alignment at *both* the representational (R) and the timing (S/E) levels.

Cognate facilitation is the same alignment phenomenon one level down. Overlapping phonological and lexical features at R raise the resting activation of a shared bundle, so the bundle is assembled faster and stabilized with a smaller Commit burst. Treating cognate facilitation, shared-gender facilitation, and shared word-order facilitation as a single alignment mechanism (differing only in which features overlap) is what allows ROSE to collapse the lexical/syntactic distinction that might otherwise be problematic here.

### *4.4. L2- L1 Effects*

Because motif entrenchment is experience-dependent and continuously updated, the model is not committed to the L1 permanently holding the highest-activation motifs. Under sustained immersion, the resting activation and gating reliability of L2 motifs can rise to the point where they win Commit and gating competitions during L1 production – yielding reverse (L2 to L1) transfer. The prediction is symmetric to the L1 to L2 case: the direction of transfer follows the current, usage-weighted attractor landscape rather than a fixed L1 advantage. Empirically, this predicts that the same

Commit and PAC-alignment signatures should index non-native-like L1 output in dominant-L2 bilinguals, with the roles of the two trajectories exchanged. This makes bidirectionality a quantitative consequence of the model rather than a separate stipulation.

## 5. Analyzing the V2 Production

In this section, we will develop these hypotheses in relation to the specific case of word order transfer in L2 Danish. As reviewed above, learners with a non-V2 background are more likely to produce non-native-like V3 sequences. In Danish, the target order is verb-second: a clause-initial adverbial is followed by the finite verb, then the subject – *Nu bor jeg i Danmark* (literally, 'Now live I in Denmark'). The non-native-like, L1-English-congruent output places the subject before the verb – *Nu jeg bor i Danmark* ('Now I live…'), i.e., V3. Søby and Kristensen (2025) found V3 rates of 26.9% for non-V2-L1 learners and 9.5% for V2-L1 learners. The study further identified clause-initial constituent length and multi-word subjects as distinct predictors of target V2 accuracy. How might we account for these findings?

### *5.1. Explaining the Non-native-like V3 Production: Sequencing Failure*

In native Danish production, θ-band S rhythms create a thematic verbal slot immediately following the first constituent, opening up subsequent frontotemporal regions for saturation by PAC with local feature bundles. However, for a learner with a non-V2 background (e.g., English), the entrenched L1 θ-band rhythm gates the Subject-related features and DM in that temporal position.

The non-native-like V3 production, therefore, is a sequencing failure (Pathway B, above). The learner effectively retrieves the correct lexico-semantic motifs (they know the relevant phonological, formal and semantic features for 'I' and 'went') but fails to suppress the L1 low-frequency prior set to coordinate these features in a manner distinct from the correct morphosyntactic sequence. This L1 rhythm keeps the Subject gating window open after the initial adverb, forcing the system to slot the subject into the position reserved for the verb. We assume that once these erroneous morphosyntactic linearization instructions are received by relevant frontal sites (e.g., pars triangularis, posterior middle frontal gyrus; Hickok, 2025; Yeaton, 2025) the transfer becomes not a failure of representational knowledge but a failure of temporal control.

Crucially, Søby and Kristensen (2025) found that non-target V3 rates increased with the number of words in the clause-initial constituent. This pattern is difficult to explain via static rules alone, but is a natural prediction of Pathway A (Subspace Competition). A longer or structurally more demanding pre-verbal constituent requires the system to maintain a specific neural state for a longer duration before the verb can be triggered. This maintenance relies on sustained β-band synchronization (Chen et al., 2026; Lundqvist et al., 2024). The longer this state must be held, the more opportunities there are for the entrenched L1 attractor to destabilize the L2 plan. As the cognitive load increases, the β power required to maintain the L2 subspace against the L1 subspace becomes insufficient, leading to a 'collapse' back into the native SVO order. Søby and Kristensen (2025) also observe a separate effect of subject length, with more V3 production for multi-word than for single-word subjects. This finding does not provide the same direct evidence, as subject length does not necessarily extend the duration between the initial constituent and the target finite verb. Because most one-word subjects in the study were personal pronouns, the effect may instead reflect broader planning load, differences in subject accessibility, or stronger entrenchment of V2 trajectories with pronominal subjects. The available data do not distinguish these possibilities. Under ROSE, these factors could increase competition during bundle assembly or reduce β-mediated stabilization of the less familiar L2-compatible trajectory.

Because the present proposal relies on properties such as a phase-grid misalignment mechanism, the critical testable observables are not mean amplitudes of time-locked ERPs, but trial-wise measures of rhythmic timing: PAC strength and burst rate/amplitude within defined planning windows. Accordingly, a mechanistic account of CLI in production should be framed in terms of induced oscillatory dynamics; how slow phase codes gate (Chen et al., 2026), and how coupling strength stabilizes (Watrous et al., 2015), competing L1 and L2 plans across time.

## 6. Limitations and Future Directions

Several limitations bound the present proposal. First, its central empirical risk is the phase-gating claim of Pathway B: existing oscillatory evidence in language most robustly supports low-frequency *tracking* and *segmentation* of structure, and while phase-dependent excitability and phase coding are established more generally, direct evidence that a specific delta/theta phase gates a specific morphosyntactic operation during production remains to be obtained. We have framed Hypothesis 2 so that this is the decisive test. Second, our worked case is a single structure (V2/V3) in one language pair; decisive evaluation requires extension to typologically diverse cases, especially agglutinative and polysynthetic languages where the linearized units are sub-lexical, and to facilitation and reverse-transfer cases beyond the illustrative examples given. Third, our account treats proficiency as continuous carving of motifs and gating reliability, but it does not yet specify the learning dynamics by which input reshapes the attractor landscape. Modeling that trajectory – including the transition from incomplete-acquisition failures to competition failures – is left to future work. Fourth, because the proposed dynamics are domain-general, isolating the language-specific contribution will require the conflict-versus-load controls described above; absent them, the signatures cannot by themselves adjudicate between structural competition and generic production difficulty. Beyond these methodological limitations, a further empirical direction concerns the analysis of non-native-like productions and comprehension outcomes. Experimental studies often include tasks targeting morphosyntactic phenomena such as gender-agreement violations. However, electrophysiological analyses may focus on familiar items and trials with correct responses, characterizing the neural response to successfully detected violations rather than the neural activity accompanying erroneous judgments (von Grebmer zu Wolfsthurn et al., 2021). Re-examining these non-target responses could provide insight into whether errors reflect lexical uncertainty, L1-based transfer, or both. Extending this approach to naturalistic speech production and comprehension settings would make it possible to compare learners at different proficiency levels and identify robust signatures of transfer across contexts.

## 7. Conclusion

One general way to summarize our multiple threads of argument in this article would be as follows: Under ROSE, L1-L2 competition can be recast as phase-space competition. Candidate L1 and L2 assemblies at R/O are simultaneously accessible to structure-building scales (S/E) but vie for privileged nesting within the same δ/θ cycle, with the selected plan reflected in the strongest coupling to the slow pacemaker and temporal centering on the cycle's excitability trough. While our explanation for syntactic interference effects may appear somewhat departed from more commonly discussed neural signatures, it is founded on similar methodological and philosophical assumptions used to explain the closely related topic of language deficits in non-neurotypical populations. Indeed, disruptions to or differences in phase-amplitude coupling dynamics have been highly correlated with syntactic and semantic parsing deficits (Meyer et al., 2021). Given this growing body of research, we expect that the prospects for using neural dynamics to enhance general theories of language dysfunction are strong. This motivates the broader possibility that CLI in bilingual production may constitute a tractable case of language dysfunction in which mechanistic biomarkers can be sought at the level of rhythmic coordination rather than solely time-locked potentials.

Importantly, our proposal is intended to complement rather than replace functional models of bilingual production such as MOGUL. At the functional level, these frameworks provide principled accounts of representational competition, control demands, and graded proficiency effects. Models of neural dynamics can add an implementational layer by specifying candidate neural mechanisms through which selection and control could be physically realized; namely, β-band stabilization of committed plans and δ/θ phase-coded gating of sequential recruitment. A natural direction for future work is therefore to further map the constructs of functional competition (e.g., degree of co-activation, inhibitory demand, similarity-based interference) onto quantifiable oscillatory observables. This translation would allow computational theories of CLI to generate more mechanistically constrained predictions about *when* and *how* transfer patterns arise during planning, and to exploit dependent measures beyond accuracy, reaction time, and ERP amplitude alone.

We also note that other theories of the neural dynamics of syntax may offer distinct predictions for CLI, and we encourage other researchers to develop more robust alignments here. It is beyond the scope of our present goals to outline potential predictions for other models that we are less familiar with and which have not been developed with language dysfunction in mind, but it would be of interest to explore how other models represent various kinds of syntactic interference and language dysfunction. We adopt ROSE not as one interchangeable option among many but because its explicit, phase-coded linking hypotheses are what let us turn functional descriptions of CLI into falsifiable production signatures, and because the same framework already underwrites a body of work characterizing language and speech deficits as 'neural dysrhythmias' (Benítez-Burraco & Murphy, 2016, 2019; Murphy & Benítez-Burraco, 2017, 2018; Jiménez-Bravo et al., 2017; Benítez-Burraco et al., 2023), akin to how we have presently discussed CLI patterns (see also Hirano et al., 2008, 2020; Kirihara et al., 2012). Our pointers to other dynamical frameworks are therefore an invitation to comparison, not a retreat from this commitment.

Certain of these competing dynamical frameworks also make different commitments about what oscillations index. As mentioned above, hierarchical tracking accounts emphasize low-frequency structure tracking as the primary signal of constituent building, often remaining agnostic about how symbolic operations are discretized into algorithmic actions (Ding, 2020, 2023; Ding et al., 2016; for related accounts and discussion, see Doelling et al., 2014; Ghitza, 2017; Meyer, 2018). Compositional neural architecture proposals emphasize structured representations and their integration (Coopmans et al., 2023; Martin, 2020), but typically do not specify a concrete control scheme that could directly predict non-native-like word-order outputs. In contrast, ROSE explicitly links symbolic operations to phase-coded control: different phases within a slow cycle can license distinct parsing actions (see Murphy, 2025, Figure 5) (e.g., lexical retrieval vs. constituent reduction), yielding concrete, falsifiable signatures. We have argued here that these features are precisely what bilingual production research requires, because CLI outputs are naturally understood as action-selection or gating errors under co-activation (Smith & Truscott, 2014).

Overall, our proposal to extend the study of non-native-like bilingual production beyond traditional timing signatures (e.g., ERPs) to include oscillatory signatures captured under models such as ROSE, offers implementational-level constraints and new dependent measures for theories of bilingualism.